\documentclass{article} 
\usepackage{iclr2024_conference,times}

\usepackage{amsmath,amsfonts,bm}

\def\eqref#1{equation~\ref{#1}}

\def\1{\bm{1}}

\DeclareMathAlphabet{\mathsfit}{\encodingdefault}{\sfdefault}{m}{sl}
\SetMathAlphabet{\mathsfit}{bold}{\encodingdefault}{\sfdefault}{bx}{n}

\usepackage{setspace}
\usepackage{url}
\usepackage{booktabs}       
\usepackage{amsfonts}       
\usepackage{amsmath}
\usepackage{amsthm}
\usepackage{amssymb}
\usepackage{subfigure}
\usepackage{nicefrac}       
\usepackage{microtype}      
\usepackage{xcolor}         
\usepackage{graphicx}

\usepackage[normalem]{ulem}
\usepackage{algorithm}
\usepackage[noend]{algpseudocode}
\usepackage{xspace}
\usepackage{float}
\usepackage{tikz}
\usepackage{verbatim}
\usepackage{tabularx}
\usepackage{enumitem}
\usepackage{wrapfig}
\usepackage{multicol}
\usepackage[most]{tcolorbox}
\usepackage[framemethod=tikz]{mdframed}
\usepackage{subcaption}
\usepackage{caption}
\usepackage{multirow}
\usepackage[utf8]{inputenc}

\usepackage{pifont}

\usepackage{listings}
\usepackage{color}
\usepackage{xcolor}

\usepackage{float}
\usepackage{graphicx}

\definecolor{isarblue}{HTML}{006699}
\definecolor{isarfaintblue}{rgb}{0.0, 0.75, 1.0}
\definecolor{isargreen}{HTML}{009966}
\definecolor{red}{HTML}{990000}
\definecolor{patriarch}{rgb}{0.5, 0.0, 0.5}

\definecolor{citecolor}{HTML}{2980b9}
\definecolor{linkcolor}{HTML}{c0392b}
\usepackage[pagebackref=true,breaklinks=true,colorlinks,bookmarks=false,citecolor=citecolor,linkcolor=linkcolor]{hyperref}

\lstdefinelanguage{isabelle}{%
    keywords=[1]{type_synonym,datatype,fun,abbreviation,definition,proof,lemma,theorem,qed,corollary,have,hence,also,finally,ultimately,moreover,using,\{},
    keywordstyle=[1]\bfseries\color{isarblue},
    keywords=[2]{where,assumes,shows,fixes,and},
    keywordstyle=[2]\bfseries\color{isargreen},
    keywords=[3]{if,then,else,case,SOME,let,in,O},
    keywordstyle=[3]\color{isarblue},
    keywords=[4]{ATP},
    keywordstyle=[4]\it\color{patriarch},
    keywords=[5]{show,assume,obtain},
    keywordstyle=[5]\bfseries\color{isarfaintblue},
}

\lstdefinestyle{isabelle}{%
  language=isabelle,
  escapeinside={\&}{&},
  columns=fixed,
  extendedchars,
  basewidth={0.5em,0.45em},
  basicstyle=\singlespacing\ttfamily\small,
  mathescape,
  morecomment=[s][\bfseries\color{red}]{(*}{*)},
  morecomment=[l][\bfseries]{####},
  breaklines=true,
}

\definecolor{mybrown}{RGB}{128,64,0}
\gdef\Sepline{%
  \par\noindent\makebox[\linewidth][l]{%
  \hspace*{-\mdflength{innerleftmargin}}%
   \tikz\draw[thick,dashed,gray!60] (0,0) --%
        (\textwidth+\the\mdflength{innerleftmargin}+\the\mdflength{innerrightmargin},0);
  }\par\nobreak}

\title{LEGO-Prover: Neural Theorem Proving with Growing Libraries}

\author{Haiming Wang\textsuperscript{\rm 1}\footnotemark[1]\quad Huajian Xin\textsuperscript{\rm 1}\thanks{~~These authors contributed equally.}\quad  Chuanyang Zheng\textsuperscript{\rm 3}\quad Lin Li\textsuperscript{\rm 6}\quad Zhengying Liu\textsuperscript{\rm 2}\thanks{~~Corresponding authors.}\quad \\\textbf{Qingxing Cao\textsuperscript{\rm 1}\quad
Yinya Huang\textsuperscript{\rm 4}\quad Jing Xiong\textsuperscript{\rm 1}\quad Han Shi\textsuperscript{\rm 2}\quad Enze Xie\textsuperscript{\rm 2}\quad Jian Yin\textsuperscript{\rm 1}\footnotemark[2]\quad}\\
\textbf{Zhenguo Li\textsuperscript{\rm 2}\quad Heng Liao\textsuperscript{\rm 6}\quad 
  Xiaodan Liang\textsuperscript{\rm 1, 5}\footnotemark[2]\quad}\\
 $^1$Sun Yat-sen University\quad 
 $^2$Huawei Noah’s Ark Lab\quad 
 $^3$The Chinese University of Hong Kong\\
 $^4$City University of Hong Kong\quad 
 $^5$MBZUAI\quad $^6$Huawei HiSilicon\quad \\
 \small\{wanghm39, xinhj, caoqx, xiongj69, issjyin\}@mail2.sysu.edu.cn, cyzheng21@cse.cuhk.edu.hk, \\
 \small\{liuzhengying2, xie.enze, shi.han, lilin29, Li.Zhenguo\}@huawei.com, liao.heng@hisilicon.com,\\
 \small yinya.huang@hotmail.com, 
 xdliang328@gmail.com
 }

\begin{document}

\maketitle

\begin{abstract}

Despite the success of large language models (LLMs), the task of theorem proving still remains one of the hardest reasoning tasks that is far from being fully solved. Prior methods using language models have demonstrated promising results, but they still struggle to prove even middle school level theorems. One common limitation of these methods is that they assume a fixed theorem library during the whole theorem proving process. However, as we all know, creating new useful theorems or even new theories is not only helpful but crucial and necessary for advancing mathematics and proving harder and deeper results.
In this work, we present LEGO-Prover, which employs a growing skill library containing verified lemmas as skills to augment the capability of LLMs used in theorem proving. By constructing the proof modularly, LEGO-Prover enables LLMs to utilize existing skills retrieved from the library and to create new skills during the proving process. These skills are further evolved (by prompting an LLM) to enrich the library on another scale. Modular and reusable skills are constantly added to the library to enable tackling increasingly intricate mathematical problems. Moreover, the learned library further bridges the gap between human proofs and formal proofs by making it easier to impute missing steps. LEGO-Prover advances the state-of-the-art pass rate on miniF2F-valid (48.0\% to 57.0\%) and miniF2F-test (45.5\% to 50.0\%). During the proving process, LEGO-Prover also manages to generate over 20,000 skills (theorems/lemmas) and adds them to the growing library. Our ablation study indicates that these newly added skills are indeed helpful for proving theorems, resulting in an improvement from a success rate of 47.1\% to 50.4\%. We also release our code and all the generated skills.
\footnote{~~Code and data can be found at \url{https://github.com/wiio12/LEGO-Prover}.}

\end{abstract}

\section{Introduction}

The automation of formal reasoning tasks, such as theorem proving and mathematical proof formalization, represents a formidable challenge and an active area of research within the domain of artificial intelligence \citep{gpt-f,pact,jiang2022thor,wholeproof,BansalLRSW19,lample_hypertree_2022,dsp,jianglisa,zhao2023decomposing,yang2023leandojo,wang2023dt,liu2023fimo}. 
The process of formalizing mathematical proofs typically relies on human experts to transcribe intricate mathematical concepts into structured formal languages verifiable by interactive theorem prover like Lean \citep{de_moura_lean_2015} or Isabelle \citep{paulson_isabelle_1994}. This process, while robust, is often labor-intensive and demands a high level of expertise. 

\begin{figure}
   \centering
   \vspace{-9mm}
   \includegraphics[width=\textwidth]{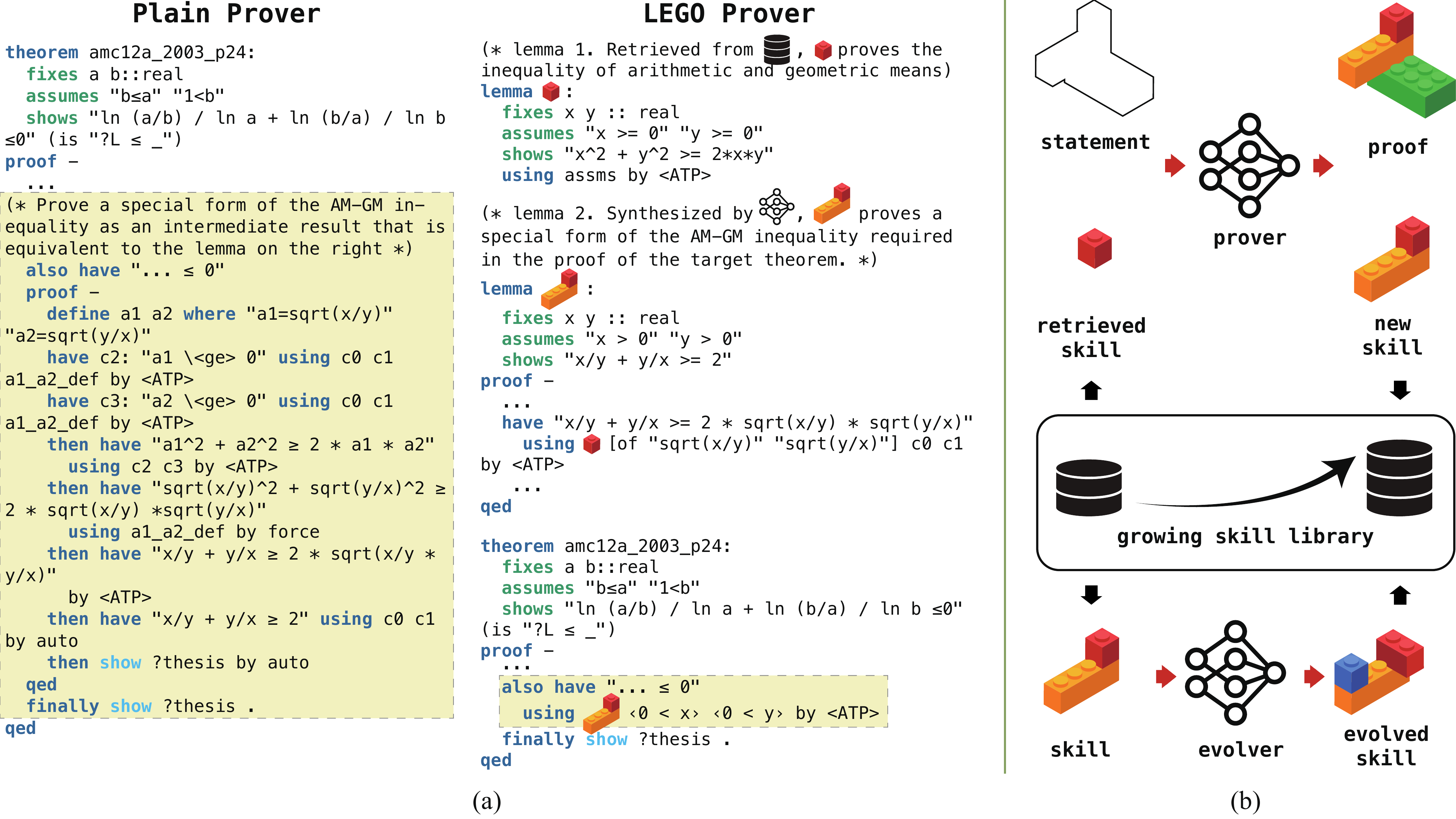}
   
  \caption{\small\textbf{(a) Comparison between plain provers and LEGO-Prover.} Unlike plain provers that prove the theorem sequentially, LEGO-Prover constructs the proof in a modular paradigm. Useful lemmas can be directly retrieved from the skill library and used as part of the proof. A newly constructed lemma can also be added to the skill library to help the proof of other theorems. \textbf{(b) Overall framework of our proposed LEGO-Prover.} LEGO-Prover contains the prover that proves the theorem modularly using the retrieved skill and the evolver that transforms the skill for reusability and generalizability. These two components are bridged by the growing skill library. }
  \vspace{-3mm}
  \label{fig:idea}
  \vspace{-1mm}
\end{figure}

In the past few years, large language models (LLMs) have emerged as a promising avenue, with their capacity to process and produce human-like text, opening doors to the idea of LLM-based neural theorem proving. Specifically, two predominant paradigms have been extensively explored in neural theorem proving. One stream of work involves step-by-step proof generation~\citep{gpt-f,pact,polu_formal_2022,lample_hypertree_2022,wang2023dt,yang2023leandojo,jiang2022thor}, where fine-tuned models provide single-step proof actions coupled with search algorithms to find the complete proofs. Another paradigm leverages the coding capabilities of LLM to construct entire proofs in a single decoding process \citep{dsp,zhao2023decomposing,wholeproof}. As shown in Fig. \ref{fig:idea}(a) left, these approaches share common proving strategies that synthesize the proof sequentially, with each step building upon the previous proof step, and stocking all the proof code into one large proof block. We denoted these approaches as \emph{plain provers} since they generate the whole proof directly.
Despite their promising results, plain provers still have several shortcomings. On one hand, plain provers attempt to prove theorems using static LLMs independently, while different problems can usually provide some insights into others. In other words, different problems may share the same lemmas while plain provers cannot utilize the proved lemmas once again even though it has been proved. 
On the other hand, even though plain provers can generate short-chain proofs with the help of advanced large language models like ChatGPT or GPT-4 \citep{gpt4} , it usually fails when it comes to long-chain proofs due to the reasoning difficulty. 

To overcome the shortcomings of plain provers and inspired by the modularity of LEGO building blocks, we present \textit{LEGO-Prover}, a novel approach designed to prove theorems in a block-by-block manner backed by a growing skill library. As shown in Fig. \ref{fig:idea}(a) right, LEGO-Prover tackles the problem of proving a theorem by first proving the sub-goal lemmas and then finalizing the problem using these lemmas. These lemmas can be retrieved from the skill library or newly constructed during the proving process. Specifically, Fig. \ref{fig:idea}(b) shows the overall process of LEGO-Prover, containing a prover and an evolver, which are bridged by the growing skill library. The prover takes the problem's formal statement as input and retrieves skills to prompt the LLM in generating the modular proof, with the generated lemmas accumulated into the skill library. However, lemmas created by the prover are often problem-specific with low reusability. Thus, LEGO-Prover incorporates an evolver that transforms the skills in the library for better generality, reusability, and complexity of the skills. The evolved new skills will also be verified and added back to the skill library. 

We conduct extensive experiments on the popular theorem-proving dataset miniF2F \citep{zheng_minif2f_2021} to validate the effectiveness of our proposed approach. LEGO-Prover significantly outperforms previous approaches, achieving a pass rate of 57.0\% and 50.0\% on the miniF2F valid and test datasets, respectively. With a 6.75\% absolute improvement on average over the previous state-of-the-art methods. In addition, our case study reveals that LLMs prove theorems modularly akin to LEGO block assembly, utilizing the retrieved skill by directly copying or using as a referee to construct the proof. Moreover, the learned skill library contains 22532 skills encompassing many useful high-level lemmas broadly applicable to various problems, as is shown in our case study and ablation study.

\section{Related works}

\textbf{Machine learning for formal mathematics.} 
Modern formal mathematics environments often center around Interactive Theorem Provers (ITPs) like Lean~\citep{de_moura_lean_2015}, Isabelle~\citep{paulson_isabelle_1994}, Metamath~\citep{megill_metamath_2019} and Coq~\citep{barras_coq_1997}. These ITPs often include specific formal languages, accompanied formal verifiers, and automated provers like Sledgehammer. ITPs provide machine-human interactive interfaces, which gives verification results during formal proof construction for specific theorems and human coder can correct errors or continue to fill gaps in proofs under the guidance of error messages and local proof states, respectively. 



Research leveraging machine learning techniques atop these formal mathematics environments generally falls into two predominant paradigms. The first focuses on proof search strategies and premise selection, epitomized by GPT-f~\citep{gpt-f}, where a language model advises single-step actions based on the current proving state, and the tree search finds a sequence of correct steps using actions given by the language model. The follow-up works PACT~\citep{pact} and Expert Iteration~\citep{polu_formal_2022} incorporate supplemental pre-training tasks like theorem naming to enhance the policy model's reasoning ability. HTPS~\citep{lample_hypertree_2022} applies Monte-Carlo tree search coupled with online training to optimize the exploration of the proof space. DT-Solver~\citep{wang2023dt} enhances search efficiency by dynamically adjusting search budgets to accommodate varying levels of state complexity. Thor~\citep{jiang2022thor} blends traditional Automated Theorem Provers (ATPs) with neural policy models to prove theorems in a neural-symbolic manner. Magnushammer~\citep{mikula2023magnushammer} augments Thor's performance by integrating premise selection, thereby boosting the performance of rule-based ATPs.

\textbf{Autoformalization.} 
Our LEGO-Prover aligns closely with the second paradigm in machine learning for formal mathematics, which leverages the capabilities of large language models (LLMs) for the formalization of mathematical proofs. Several notable works have paved the way in this domain. For instance, \cite{wang2018first} delved into both supervised and unsupervised translation techniques for auto-formalization tasks. \cite{wu2022autoformalization} makes its first attempt at employing a large language model to translate natural language mathematical problems into formal theorem statements. Building on this, Draft, sketch, and proof \cite{dsp} develops a three-step approach that aims to fully formalize proofs, using natural language as guidance. \cite{wholeproof} goes a step further by generating complete proofs in a single pass and introducing a proof repair model to enhance the theorem-proving capabilities. \cite{zhao2023decomposing} advances \cite{dsp} by incorporating cross-verified informal proofs to better inform the generation of formal proof sketches. Despite their contributions, none of the aforementioned methods have succeeded in establishing a learning paradigm that incrementally formalizes increasingly complex problems via a growing skill library, a gap that our work seeks to fill.

\textbf{Skill-based agents.} 
LEGO-Prover is also related to trending AI agents powered by large language models like GPT-3.5 and GPT-4~\citep{shen2023hugginggpt,park2023generative,wang2023unleashing,zahedi2021humanai}. These AI agents are characterized by their ability to perform complex tasks through a combination of task planning, logical reasoning, and skill accumulation. For example, Voyager~\citep{wang2023voyager} creates an AI agent capable of autonomously playing Minecraft. It has a dynamic growing skill library that empowers the in-game character to tackle increasingly intricate tasks. Similarly, \citep{cai2023large} showcases the ability to generate reusable Python tools and documentation, which can then be leveraged by weaker GPT models, increasing the ability of the model. 
\section{Method}

\begin{figure}[]
\vspace{-4mm}
   \centering
   \includegraphics[width=0.97\textwidth]{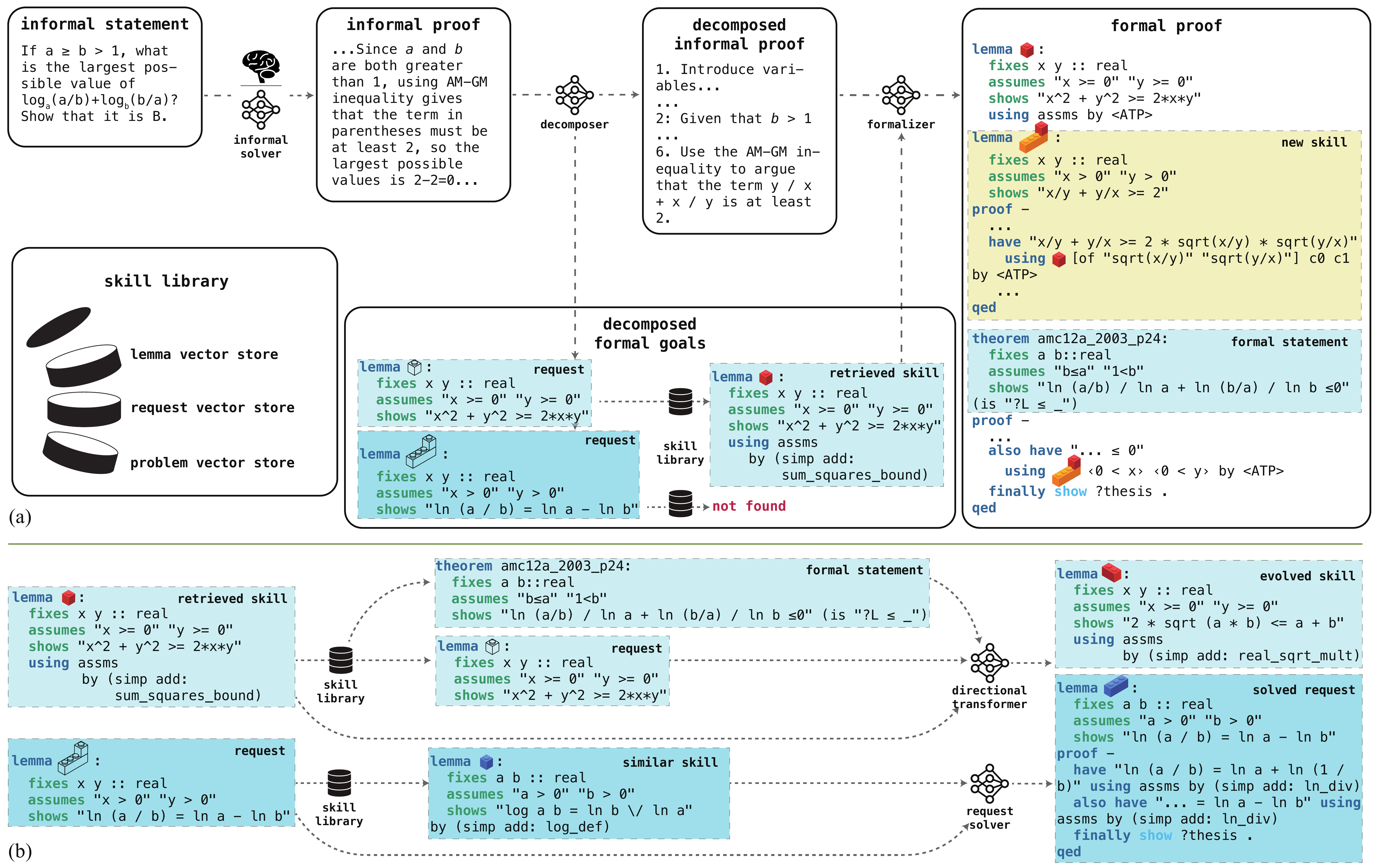}
  \caption{\small\textbf{Overview of LEGO-Prover.} (a) The prover takes three consecutive steps to prove the problem. The informal solver produces an informal solution for the problem statement, followed by using the decomposer to decompose it into step-by-step proof and propose useful sub-goals in the form of lemma statements. We use these lemma statements to retrieve useful lemma from the skill library. The formalizer constructs the final proof with the help of decomposed informal proof and retrieved lemma in a block-by-block manner. We extract newly constructed lemmas from the final proof and add them to the lemma vector store. (b) The evolver contains two evolution approaches, the directional transformer transforms the existing skill into a more general and reusable form in four predefined directions. The request solver directly solves the requested subgoal proposed by the prover. Newly generated and formally verifed lemmas from the evolver will be added to the lemma vector store.}
  \label{fig:main_dts}
\vspace{-4mm}
\end{figure}

In this section, we introduce the detailed implementations of our proposed LEGO-Prover. Following the setting of \cite{dsp}, we assume that each theorem is equipped with an informal statement, a human-written informal proof, and a formal statement defining the problem. As illustrated in Fig. \ref{fig:main_dts}, LEGO-Prover consists of two main components: the prover and the evolver. The prover decomposes the problem into possible subgoal lemmas and proves the problem in a block-by-block style, aided by the skill library. The evolver refines skills in the skill library for enhanced diversity, generality, and reusability, and also resolves decomposed sub-goals from the prover to create new skills.  These components are linked via a skill library housing lemmas and requests. In the following sections, we will detail introduce the skill library, the prover, and the evolver.

\subsection{Skill library}
The skill library contains various vector stores that are optimized for retrieval. Every vector store maintains its data in pairs consisting of documents and their corresponding embeddings, encoded with an embedding language model\footnote{~~Practically, ChromaDB serves as our vector store, coupled with the OpenAI's text-davinci-ada embedding model.}. Upon the receipt of a query, the vector store employs the embedding model to encode the query and leverages the k-NN algorithm to retrieve relevant documents stored within. The skill library used in LEGO-Prover is comprised of three distinct vector stores. 1) The lemma vector store contains the Isabelle verified lemmas, encompassing both the lemma’s statement and its proof. This is the core of the skill library and it facilitates the perpetual enhancement of LLMs' capabilities in constructing increasingly intricate theorems. For representation simplicity, the notion of lemmas in the lemma vector store and skills in the skill library are use interchangeably in this paper. 2) The request vector stores preserve the lemma statements proposed by the decomposer. These requests are crucial to the success of LEGO-Prover, their works as an in-depth reasoned query for retrieving the useful skill for the prover, as well as possible complete lemmas when they are solved by the evolver. 3) The problem vector store houses the formal statements in the miniF2F dataset. The evolver utilizes these problem statements as heuristics to guide the LLMs in generating more beneficial new lemmas.

\subsection{Prover}
As illustrated in Fig. \ref{fig:main_dts} (a), the prover employs a three-step process to construct the proof. 
Initially, an informal solver is deployed to draft a solution in natural language corresponding to the informal statement. Akin to \citep{dsp}, LEGO-Prover experiments the use of ground truth human-written proofs as alternatives to model-generated proofs.
After obtaining the informal proof, LEGO-Prover constructs the formal proof using the decomposer and the formalizer sequentially, which we detailed in the following.

\textbf{Decomposer.} The decomposer aims to decompose the formalization tasks, which transform the informal proof into the decomposed step-by-step informal proof as well as decompose the problem into formal goals. A concrete example of the decomposer is shown in Fig. \ref{fig:decomposer} in Appendix. \ref{app:prompt_example}. 
Specifically, the decomposer prompts the LLM to refine informal proofs, producing step-by-step informal proof that more closely aligns with the structure of the actual Isabelle proof code. 
We posit that this alignment is crucial as it considerably reduces the complexity encountered during the formalization process. Concurrently, the decomposer tasks the LLM with generating requests: some potential lemma or theorem statements that could be useful in addressing the given problem.
Each request is composed of a chain-of-thought reasoning on what kind of lemma is required for solving the problem followed by the formal statement of the lemma. Subsequently, LEGO-Prover put these requests into the request vector store. 

\textbf{Formalizer.} 
The process of formalization involves translating an informal proof into Isabelle's proof sketches, as depicted in Fig. \ref{fig:formalization} (refer to Appendix. \ref{app:prompt_example}). 
In addition to the problem statement, the refined informal proof, and the formal statement, the formalizer is designed to incorporate useful lemmas retrieved from the lemma vector stores as part of the input. 
The formalizer employs the proposed request originating from the decomposer and the formal statement of the problem as query texts and, in total, retrieves $n_f$ skills. Upon collecting all the necessary input, the LLM is tasked to provide the proof code. Unlike the setting in \cite{dsp} and \cite{zhao2023decomposing}, we prompt the LLM to construct the complete content of the source file in Isabelle. 
This may encompass the requisite imports, definitions, or lemmas before the elucidation of the main problem to be proven. Consequently, the LLM possesses the capability to generate or apply useful lemmas before embarking on the resolution of the problem. Empirical evaluations demonstrate that our model exhibits a more modular problem-solving approach compared to established baseline methods. This modularity facilitates recycling smaller lemma components, thereby enhancing the LLM's capability to tackle progressively intricate problems.


After obtaining the formalized proof code, LEGO-Prover employs the Isabelle theorem prover to verify the correctness of the provided proof code. In instances where a particular proof tactic (such as "by ...") falls short of proving the given conjecture, we resort to 11 heuristic tactics alongside the sledgehammer method to facilitate an auto-correction. The heuristic selection we employ is consistent with those presented in \cite{dsp}.
After verifying the code, all validated lemmas or theorems are added to the skill vector store, while any failed lemmas' statement is added to the request vector store. We consider a formalized proof valid if and only if (a) the proof does not contain "cheating" keywords (sorry or oops) that exit a proof without completing it. (b) Isabelle can verify the proof code containing the corresponding formal statement. 


\subsection{Evolver}
The lemmas extracted from the prover are mostly problem-specific, rendering them non-reusable with limited applicability. And the number of these lemmas is also very limited. The objective of the evolver is to create or refine these skills, enhancing their reusability and expanding their functional coverage.
As shown in Fig. \ref{fig:main_dts} (b), the evolver contains two functionalities: the directional transformer transforms the current skill and the request solver directly solves the request proposed by the prover to create new lemmas. We detail each in the following.

\textbf{Directional transformer.}
The objective of the directional transformer is to facilitate the evolution of a skill along various predefined trajectories, thereby augmenting the reusability and usefulness of the skill. 
It is composed of four distinct trajectories: extension of dimensions, identification of key concepts, parameterization, and enhancement of complexity. 
Table. \ref{app:prompt_example} shows the detailed functionality of each evolving direction. 
Each instance of the directional transformer adheres to a unified prompt template depicted in Fig. \ref{fig:directional_evolve}. The adaptation involves substituting the core description and its in-context examples for the corresponding evolving directions.
Specifically, the directional transformer begins with randomly selecting the least evolved skill (with the least amount of time being selected to evolve). Subsequently, the transformer employs this skill to retrieve $n_d$ relevant pending problem's formal statement from the problem vector store and the relevant request's formal statement from the request vector store. Upon assembling the inputs for the LLM, the transformer arbitrarily selects one direction of evolution and prompts the LLM to generate a novel skill. 

\textbf{Request solver.} 
The request solver is designed to facilitate the development of new skills by directly addressing the sub-goals proposed by the prover. As depicted in Fig. \ref{fig:solve_request}, the process initiated by the request solver involves the random selection of a minimally solved request (with least amount of time being selected to solve the request). After this selection, this request is employed to query the lemma vector store to retrieve pertinent skills that can serve as references. Finally, the request solver prompts the LLM to generate the proof for the request. 


After obtaining the new skill (evolved lemma or solved request) generated by the LLM, the evolver uses Isabelle to verify the generated code. 
To mitigate the risk of redundancy within the skill library, a comparative strategy is conducted between the newly acquired skills and existing ones. This is accomplished by employing the \texttt{SequenceMatcher} method from the \texttt{difflib} Python library, which quantifies the level of difference between the new and existing skills. Only skills that have been verified and exhibit a difference bellowing the predetermined threshold of $0.85$ are incorporated into the skill library. 
\section{Experiments}

    
    

\subsection{Experimental setup}
\textbf{Implementation details.} 
To expedite the experimental procedure, both the prover and the evolver are executed in a multi-processing manner, maintaining a process number ratio of $3:8$ respectively. Consistent with the \citep{dsp,zhao2023decomposing}, each problem undergoes 100 attempts of proving. To maximally leverage the expanding skill library, problems are formalized through successive rounds, with each round addressing each valid/test set problem once.


For the execution of the prover and the evolver, ChatGPT is utilized as the LLM.\footnote{~~A combination of gpt-3.5-turbo, gpt-3.5-turbo-0301, gpt-3.5-turbo-0613, gpt-3.5-turbo-16k, and gpt-3.5-turbo-16k-0613 is employed, with a model being selected randomly during calls to the OpenAI API.} The temperature is consistently set at $T=0.7$ across all procedures. Within the prover, 3-shot examples are leveraged for the decomposer. Regarding the formalizer, the quantity of reference skills $n_f$ is set to $6$ for the valid set and $4$ for the test set, and paired with 2 formalization in-context examples. For the directional transformer, the number of reference problem statements $n_d$ is set to 4, supplemented by two directional transformation in-context examples. For the request solver, 3 skills are retrieved for reference.

\textbf{Dataset and evaluation.} 
For a more accurate comparison, we follow \citep{dsp,zhao2023decomposing} and adopt the miniF2F dataset \citep{zheng_minif2f_2021}. This dataset includes 488 problems sourced from high-school mathematical competitions. These problems vary in difficulty, ranging from basic algebra and number theory, originating from the MATH dataset \cite{hendrycks2021measuring}, to more challenging problems found in the American Invitational Mathematics Examination (AIME) and International Mathematical Olympiad (IMO). The problems are divided into valid and test sets, with 244 problems each. In this paper, we utilize the updated version of the miniF2F dataset from \cite{dsp}. Each question in this updated dataset contains a formal statement in Isabelle language, an informal statement, and a human-written informal proof. For interacting with the Isabelle theorem prover, we employ the PISA environment \citep{jianglisa}. PISA is a flexible Python REPL wrapper for Isabelle, capable of verifying Isabelle's code and providing information such as proof states or error messages from Isabelle.

\textbf{Baseline methods.} 
\begin{table*}[t]
\begin{center}
\vspace{-4mm}

\caption{
\small
\textbf{Proving success rates on the miniF2F dataset with Isabelle.} 
The table displays the success rates of previous works and the LEGO-Prover, The highest success rates for each set are highlighted in bold. LEGO-Prover* denotes the cumulative pass rate of the miniF2F dataset, considering the total number of problems solved using model-generated and human-written proofs. 
}
\label{tab:main_results} 
\small
\begin{tabular}{lccc}
    \toprule
    Success rate & LLM & miniF2F-valid & miniF2F-test \\
    \midrule
    \multicolumn{1}{l}{\textit{Baselines}} \\
    \midrule
    Thor~\citep{jiang2022thor} & - & $28.3\%$ & $29.9\%$ \\
    Thor + expert iteration~\citep{wu2022autoformalization} & Codex & $37.3\%$ & $35.2\%$ \\
    Draft, sketch, and Prove~\citep{dsp} & Codex & 42.6\% & 39.3\% \\
    Subgoal-Learning~\citep{zhao2023decomposing}& ChatGPT &48.0\% & 45.5\% \\
    \midrule
    \multicolumn{1}{l}{\textit{Ours (100 attempts)}} \\
    \midrule
    LEGO-Prover (model informal proof) & ChatGPT  & $52.4\%$ & $45.5\%$ \\
    LEGO-Prover (human informal proof) & ChatGPT & $55.3\%$ & $\mathbf{50.0\%}$ \\
    LEGO-Prover*  & ChatGPT & $\mathbf{57.0}\%$ & $\mathbf{50.0\%}$ \\
    \midrule
    \multicolumn{1}{l}{\textit{Ablations (50 attempts)}} \\
    \midrule
    LEGO-Prover         & ChatGPT   & $50.4\%$ & - \\
    - Skill Library   & ChatGPT      & $47.1\%$ & - \\
    
    \bottomrule
\end{tabular}
\end{center}
\vspace{-0.1in}
\end{table*}
We have included baselines that represent state-of-the-art neural theorem proving in Isabelle. Thor\citep{jiang2022thor} and Thor with expert iteration on auto-formalized data \citep{wu2022autoformalization} are works focused on proof search paradigms, which use a fine-tuned 700m language model to prove theorems. Draft, Sketch, and Prove \citep{dsp} and Subgoal-Learning \cite{zhao2023decomposing} are works that use Codex or ChatGPT to prove theorems directly.

Following the setting from \cite{dsp}, we test the LEGO-Prover with model-generated and human-written informal proofs. The model-generated informal proofs are pre-generated using GPT-4, with up to 20 informal proofs per problem. For each proving attempt, we randomly select one proof as the informal proof to feed into the decomposer procedure. For ablation, we remove the growing skill library to validate the effectiveness of the LEGO-Prover. Due to limited resources and the expense of OpenAI API calls\footnote{~~Estimated to be around 300 dollars for one experiment with 100 proof attempts}, we perform ablation only under 50 proving attempts per problem on the miniF2F validation set.


\subsection{Main result}



\begin{figure}[t]
    \centering
\setlength{\belowcaptionskip}{-0.3cm}
    \vspace{-4mm}
    \subfigure[]{\includegraphics[width=0.495\textwidth]{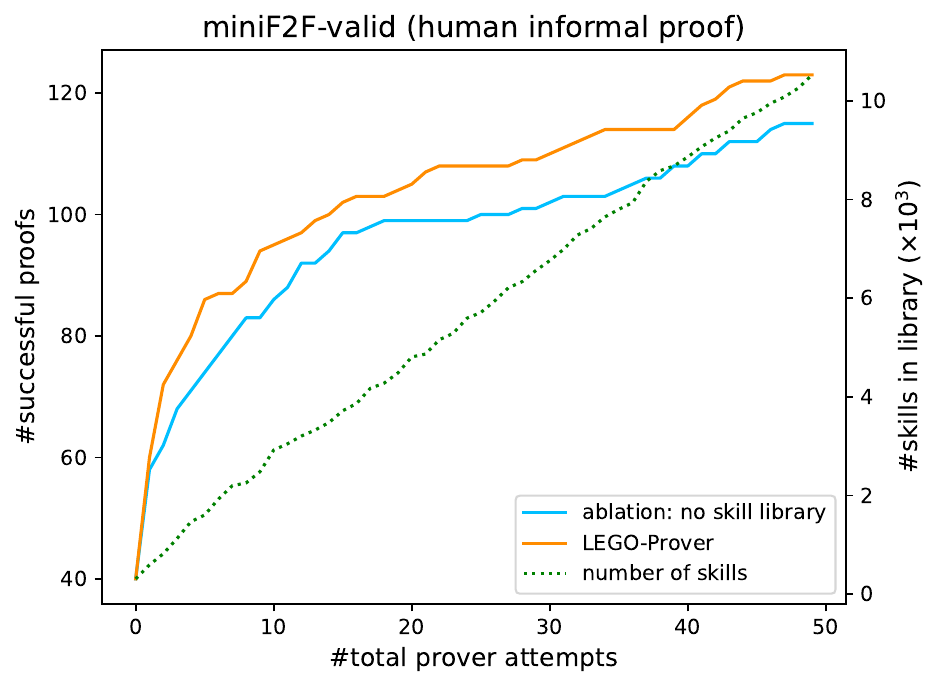}} 
    \hspace{10pt}
    \subfigure[]{\includegraphics[width=0.46\textwidth]{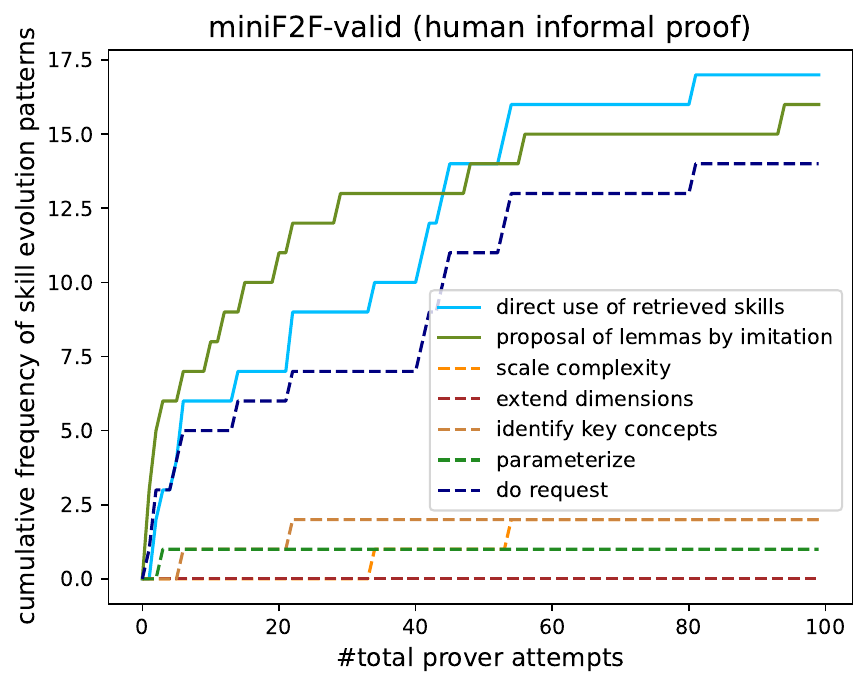}}
    \subfigure[]{\includegraphics[width=0.98\textwidth]{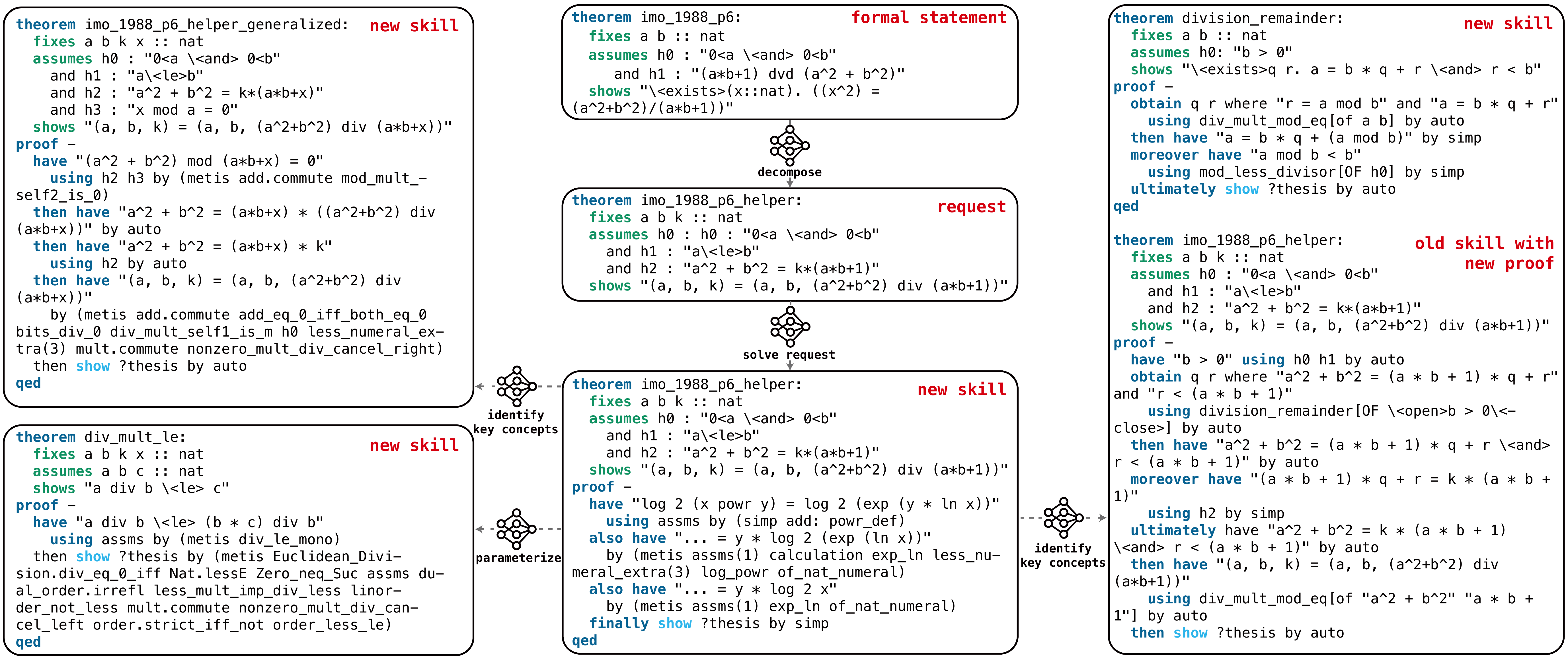}}
    \caption{\small(a) A comparison of proof success rate between LEGO-Prover with and without the growing skill library. As the number of skills grows (shown in green dotted line), the prover's performance gap between with and without the skill library becomes increasingly evident  
    (b) Distribution of skill origins in successful proofs, plotted against prover attempts and the percentage contribution of each skill origin to the total successful proofs. The solid lines show the distribution of the two major methods by which the skill is used (direct use or proposing a lemma by imitation). The dotted lines show the detailed skill origins used in the successful proofs.
    (c) A skill-evolving tree gradually generated through multiple steps for {\tt imo\_1988\_p6} conceals how a relatively large-scale skill library is produced from a few seed theorems.}
    \label{fig:results}
    \vspace{-2mm}
\end{figure}

In Table. \ref{tab:main_results}, we illustrate the proportion of successful formal proofs found on the miniF2F dataset. Thor and Thor + expert iteration displays the performance of the fine-tuned search-based proving method. The results indicate that all LLM-based methods significantly outperform the search-based methods by around 4.7\%. The efficacy of search-based proving methods is limited by the short proof steps that the policy language model generates, drastically increasing the search space and, therefore, hindering the prover from finding long proofs.

Our proposed LEGO-Prover significantly outperforms both search-based and LLM-based methods. With proofs written by humans, the LEGO-Prover improves by 7.3\% and 4.5\% over the Subgoal-Learning method on the miniF2F-valid and miniF2F-test datasets, respectively. A total of 257 out of 488 problems were solved by the LEGO-Prover with human-written proof. When replacing human-written informal proofs with model-generated informal proofs, the LEGO-Prover still achieves 52.4\% and 45.5\% pass rates on the valid set and test set, respectively, close to the results with human-written informal proofs.


\textbf{Effects of the growing skill library.} The growing skill library greatly enhances the proving ability of static LLMs like chatGPT or GPT-4. As the major contribution of LEGO-Prover, we are interested in how well it contributes to solving more problems and improving the LLMs' ability. Specifically, we remove the growing skill library (including the evolver). As shown in the Table. \ref{tab:main_results}, in 50 attempts, LEGO-Prover achieves 50.4\% on the validation set, whereas the LEGO-Prover without a skill library achieves 47.1\%. For a more intuitive representation, we further plot the trends of the number of problems solved against the number of proving attempts for both settings, shown in Fig. \ref{fig:results}(a). 
Compared to the problem solved without a growing skill library, the advantage of adding the skill library is initially minimal, as the libraries are still underdeveloped and lack useful skills. However, as the skill library expands, the gap between LEGO-Prover and the ablation method widens consistently. This outcome supports our hypothesis that the prover becomes increasingly adept at formalizing theorems as more skills are added to the skill library.


\subsection{Analysis}

\subsubsection{What does the growing skill library look like?}
Figure. \ref{fig:results}(c) illustrates an example of a skill-evolving tree in the skill library. The grown skill library is a massive forest containing numerous evolved trees like this one. The lemmas, originating from either the prover or the evolver's request solver sub-task (as the example shown in the figure), become the root nodes of these skill-evolving trees. The evolver's directional transformation generalizes these lemmas and creates child nodes. In terms of statistics, there are 22532 skills in the skill library in total, with 10.8\% of the skills originating from the prover, 38.2\% originating from the evolver's solve request sub-task, and 51.1\% originating from the evolver's directional transformation sub-tasks. Although some lemmas are trivially true or already exist in Isabelle's theorem library, LEGO-Prover generates more unique, interesting, and useful lemmas. The gap between natural and formal language is the greatest challenge in formalizing natural language mathematical proofs. A simple proving step described in natural language can result in numerous lines of code written in the formal language. However, this gap can gradually diminish as we accumulate more lemmas and theorems. To conclude, the grown skill library generated by LEGO-Prover significantly enlarges this library and further bridges the gap between informal and formal mathematical languages.

\subsubsection{How does the skill boost the ability of LLMs?}

\begin{figure}[t]
\vspace{-4mm}
\centering
\includegraphics[width=1\textwidth]{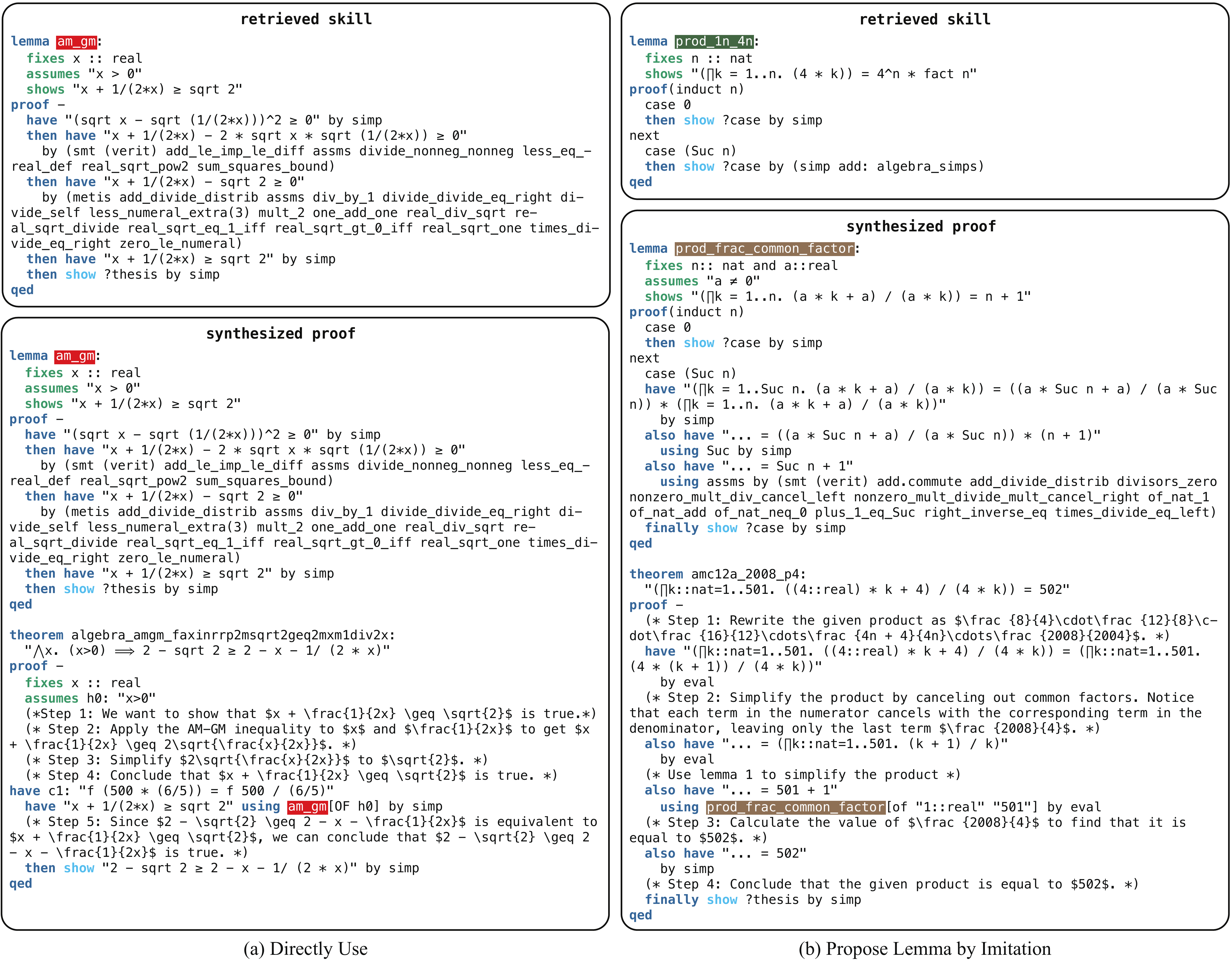}

\caption{\small\textbf{Two primary forms of utilizing the skills.} (a) \textbf{Directly use.} we retrieved the \texttt{am\_gm} skill from the skill library and provided it as a reference skill for the formalizer. The formalizer synthesizes proof by directly copying the lemma as part of the proof code and using the lemma in the main proof. (b) \textbf{Propose lemma by imitation.} The retrieved lemma \texttt{prod\_1n\_4n} serves as a useful reference for the prover to synthesize the lemma \texttt{prod\_frac\_common\_factor}. }
\label{fig:case_study}
\vspace{-15pt}
\end{figure}
To investigate closely how these learned skills can better help and boost the performance of LLM, we manually inspect the successfully proven problems in the miniF2F valid set. The conclusions are as follows:

\textbf{Skill as directly reusable blocks.} This is the most straightforward way of utilizing the generated skills. Since every skill in the skill library is verified by the Isabelle prover, LLM can directly copy the lemma code from the input prompt without fear of causing an error. As shown in Fig. \ref{fig:case_study} left, the final proof of the problem {\tt algebra\_amgm\_faxinrrp2msqrt2geq2mxm1div2x} directly copies the retrieved skill {\tt am\_gm}'s code as part of the proof and uses this lemma to help prove the problem.

\textbf{Skill as reference for solving the problem.} Many skills cannot be directly reused but are very helpful as reference examples for formalizing the main problem. As shown in Fig. \ref{fig:case_study} right, the retrieved skill examples {\tt prod\_1n\_4n} provide great clues for solving the conjecture {\tt prod\_frac\_common\_factor}. Since the provided skills are lemmas with verified steps, these steps drastically increase the accuracy of the LLM to generate the correct proof steps. 

Fig. \ref{fig:results}(b) first compares two scenarios: directly applying retrieved skills to the proofs and constructing new lemmas by imitating retrieved skills to assist in theorem proving (represented by the light blue and light green lines). It then examines the skill evolution pattern of the lemmas used in the proofs (corresponding to Fig. \ref{fig:main_dts} (b) and Table. \ref{tab:sssd}). Out of 135 problems of the miniF2F-valid dataset passing the validation of the Isabelle verifier, 24\% is completed with the aid of retrieved skills. Within this subset, 51\% of the problems directly incorporate the retrieved skills into their proofs, while the remaining 49\% formulate new lemmas that are specifically tailored to address the distinct problems at hand. Regarding the skills directly applied in the proofs, 71\% are procured by the "do requests" procedure. The skills derived through the evolution techniques of "identifying key concepts" and "scaling complexity" each contributes to 12\%, while those acquired through "parameterization" constitute 6\%. Although skill as directly reusable blocks is the most ideal usage of skills in the library, the problems solved by directly reusing the skill are not substantial. That is because many trivial problems in the dataset miniF2F can be solved trivially without requiring any skill as a reference. 

\section{Conclusions}
In this work, we introduced a new theorem-proving method, LEGO-Prover, which uses a growing skill library to continuously boost the capability of LLM for formalization. The prover utilizes refined structural informal proof and retrieved lemma to correctly formalize the proof. The evolver solves the request proposed by the prover or evolves existing skills into new skills. LEGO-Prover introduces a fundamentally different theorem proving paradigms for the community. With the previous approaches all struggling to complete the proof at once, LEGO-Prover tries to prove the theorem in a block-by-block manner, akin to divide and concur approaches. Extensive tests show that our method can indeed improve pass rates on the miniF2F dataset. Our ablation studies and detailed analysis showcase the effectiveness of each component we proposed in LEGO-Prover. 



\nocite{*}
\bibliography{iclr2024_conference}
\bibliographystyle{iclr2024_conference}

\clearpage
\appendix
\section{Appendix}

\definecolor{isarblue}{HTML}{006699}
\definecolor{isarfaintblue}{rgb}{0.0, 0.75, 1.0}
\definecolor{isargreen}{HTML}{009966}
\definecolor{red}{HTML}{990000}
\definecolor{patriarch}{rgb}{0.5, 0.0, 0.5}

\lstdefinelanguage{isabelle}{%
    keywords=[1]{type_synonym,datatype,fun,abbreviation,definition,proof,lemma,theorem,qed,corollary,have,hence,also,finally,ultimately,moreover,using,\{},
    keywordstyle=[1]\bfseries\color{isarblue},
    keywords=[2]{where,assumes,shows,fixes,and},
    keywordstyle=[2]\bfseries\color{isargreen},
    keywords=[3]{if,then,else,case,SOME,let,in,O},
    keywordstyle=[3]\color{isarblue},
    keywords=[4]{ATP},
    keywordstyle=[4]\it\color{patriarch},
    keywords=[5]{show,assume,obtain},
    keywordstyle=[5]\bfseries\color{isarfaintblue},
}

\lstdefinestyle{isabelle}{%
  language=isabelle,
  escapeinside={\&}{&},
  columns=fixed,
  extendedchars,
  basewidth={0.5em,0.45em},
  basicstyle=\singlespacing\ttfamily\tiny,
  mathescape,
  morecomment=[s][\bfseries\color{red}]{(*}{*)},
  morecomment=[l][\bfseries]{####},
  breaklines=true,
}

\definecolor{mybrown}{RGB}{128,64,0}
\gdef\Sepline{%
  \par\noindent\makebox[\linewidth][l]{%
  \hspace*{-\mdflength{innerleftmargin}}%
   \tikz\draw[thick,dashed,gray!60] (0,0) --%
        (\textwidth+\the\mdflength{innerleftmargin}+\the\mdflength{innerrightmargin},0);
  }\par\nobreak}

\subsection{Prompt examples}
\label{app:prompt_example}
In this section, we illustrate the prompts used in the LEGO-Prover. For prover, the prompt used is the decomposer (Fig. \ref{fig:decomposer}), and the formalizer (Fig. \ref{fig:formalization}). For the evolver, the prompt used is the directional evolver (Fig. \ref{fig:directional_evolve}) and request solver (Fig. \ref{fig:solve_request}). The blue line separates the LLMs' input and outputs.

For directional evolve, we list all the core statement to be replaced in the Table. \ref{tab:sssd}

\begin{table}[h]

 \centering
  \caption{The core description of individual directional evolve. The description will be replaced into the directional evolve prompt template.}
  \footnotesize
\setlength\extrarowheight{1.5pt}
\setlength{\tabcolsep}{5pt}
  \begin{tabular}{c|l}
  \toprule
    Evolve type & Description  \\
\hline
 \multirow{2}{*}{Identify key concepts} &  Determine the essential ideas, methods, or \\
    & theorems that are crucial to solving the initial problem. \\ 
\hline
\multirow{2}{*}{Parameterize} &  If the problem involves specific numbers, generalize \\ 
 &it by replacing these with variables. \\ 
 \hline
 \multirow{2}{*}{Scale complexity} & Try both simpler and more complicated versions of the \\
    & problem to see how the approach adapts. \\
    \hline
 \multirow{2}{*}{Extend dimensions} & If the problem is defined in a specific number of dimensions, \\
 & consider if it holds in more or fewer dimensions. \\    
  \bottomrule
\end{tabular}
\label{tab:sssd}
\vspace{-2mm}
\end{table}


\clearpage

\begin{figure}[H]
\begin{tcolorbox}[colback=mybrown!5!white,colframe=mybrown!75!black]
\begin{tiny}
\textbf{\small Input:\\\ \\}
\textbf{System message:}
As a mathematician and expert in Isabelle theorem prover, your task is to analyze the given theorem (including the problem's informal statement, human written informal proof, and formal statement). Provide a better structured step-by-step proof that is closer to Isabelle. and request relevant lemmas, and theorems that might help in proving this problem.
\textbf{Statement:}
If $a\geq b > 1,$ what is the largest possible value of $\log_{a}(a/b) + \log_{b}(b/a)?$  Show that it is 0.
\\
\textbf{Informal proof:} 
Using logarithmic rules, we see that  $\log_{a}a-\log_{a}b+\log_{b}b-\log_{b}a = 2-(\log_{a}b+\log_{b}a)$ $=2-(\log_{a}b+\frac {1}{\log_{a}b})$  Since $a$ and $b$ are both greater than $1$, using [[AM-GM]] gives that the term in parentheses must be at least $2$, so the largest possible values is $2-2=0$  Note that the maximum occurs when $a=b$.
\\
\textbf{Formal statement:}
\vspace{-12pt}
\begin{lstlisting}[style=isabelle]
theorem
  fixes a b::real
  assumes "b\<le>a"
    and "1<b"
  shows "ln (a/b) / ln a + ln (b/a) / ln b \<le>0" (is "?L \<le> _")
\end{lstlisting}
\vspace{-6pt}
\textbf{Structure proof:}
Step 1: Introduce variables x and y to represent the natural logarithms of a and b respectively.\\
Step 2: Given that \(b > 1\), conclude that the natural logarithm of b, \(y\), is greater than 0.\\
Step 3: Using the given assumptions, establish that \(x \ge y\). This implies \(x > 0\).\\
Step 4: Express the LHS of the main inequality in terms of x and y.\\
Step 5: Simplify the expression derived in Step 4.\\
Step 6: Demonstrate that the above expression is less than or equal to 0. The Arithmetic Mean-Geometric Mean Inequality (AM-GM) is used here to argue that the term \((y/x + x/y)\) is at least 2, hence the expression is at most 0.
\\
\textbf{Required skills:}\\
Thoughts 1: A base lemma proving am\_gm inequality that is useful for step 6\\
Code 1: \\
\vspace{-21pt}
\begin{lstlisting}[style=isabelle]
lemma am_gm:
  fixes x y :: real
  assumes "x >= 0" "y >= 0"
  shows "x^2 + y^2 >= 2*x*y"
\end{lstlisting}
\vspace{-6pt}
Thoughts 2: According to the step 6, we derived a special form of am\_gm inequality. which proofs the conjecture $x/y + y/x >= 2$ required in step 6.
\\
Code 2: \\
\vspace{-21pt}
\begin{lstlisting}[style=isabelle]
lemma am_gm_divide_form:
  fixes x y :: real
  assumes "x > 0" "y > 0"
  shows "x/y + y/x >= 2"
\end{lstlisting}
\vspace{-6pt}
[two more examples like above...]
\textbf{Statement:}
What is the least possible value of $(xy-1)^2+(x+y)^2$ for real numbers $x$ and $y$? Show that it is 1.
\\
\textbf{Informal proof:}\\
Expanding, we get that the expression is $x^2+2xy+y^2+x^2y^2-2xy+1$ or $x^2+y^2+x^2y^2+1$. By the Trivial Inequality (all squares are nonnegative) the minimum value for this is 1, which can be achieved at $x=y=0$.
\\
\textbf{Formal statement:}
\vspace{-12pt}
\begin{lstlisting}[style=isabelle]
theorem amc12a_2021_p7:
  fixes x y ::real
  shows "1 \<le> ((x * y) - 1)^2 + (x + y)^2"
  apply (auto simp:algebra_simps power2_eq_square)
\end{lstlisting}
\vspace{-6pt}
\Sepline
\textbf{ \ \\\small Output:\\\ \\}
\textbf{Structure proof:}\\
Step 1: Expand the expression \((xy-1)^2+(x+y)^2\) to obtain \(x^2+2xy+y^2+x^2y^2-2xy+1\).\\
Step 2: Simplify the expression derived in Step 1 to obtain \(x^2+y^2+x^2y^2+1\).\\
Step 3: Apply the Trivial Inequality, which states that all squares are nonnegative, to argue that the minimum value of the expression is 1.\\
Step 4: Show that the minimum value of 1 can be achieved when \(x=y=0\).\\
\textbf{Required skills:}\\
Thoughts 1: The Trivial Inequality is a key concept in this proof. Understanding and applying this inequality is crucial to show that the minimum value of the expression is 1.
\\
Code 1: \\
\vspace{-21pt}
\begin{lstlisting}[style=isabelle]
lemma trivial_inequality:
  fixes a :: real
  shows "0 \<le> a^2"
\end{lstlisting}
\vspace{-6pt}
Thoughts 2: The ability to expand and simplify algebraic expressions is important in this proof.
\\
Code 2: \\
\vspace{-21pt}
\begin{lstlisting}[style=isabelle]
lemma expand_expression:
  fixes x y :: real
  shows "(x * y - 1)^2 + (x + y)^2 = x^2 + 2 * x * y + y^2 + x^2 * y^2 - 2 * x * y + 1"
\end{lstlisting}
\vspace{-6pt}
Thoughts 3: The ability to substitute values into an expression and evaluate it is necessary to show that the minimum value of 1 can be achieved when $x=y=0$.
\\
Code 3: \\
\vspace{-21pt}
\begin{lstlisting}[style=isabelle]
lemma substitute_values:
  fixes x y :: real
  assumes "x = 0" "y = 0"
  shows "(x * y - 1)^2 + (x + y)^2 = 1"
\end{lstlisting}
\vspace{-6pt}

\end{tiny}
\end{tcolorbox}
\caption{Prompt example for decomposer. }
\label{fig:decomposer}
\end{figure}

\begin{figure}[H]
\begin{tcolorbox}[colback=mybrown!5!white,colframe=mybrown!75!black]
\begin{tiny}
\textbf{\small Input:\\\ \\}
\textbf{System message:}
As a mathematician familiar with Isabelle, your task is to provide formal proof in response to a given problem statement.
Your proof should be structured and clearly written, meeting the following criteria:
\\- It can be verified by Isabelle.
\\- Each step of the proof should be explained in detail using comments enclosed in "(*" and "*)".
\\- The explanation for each step should be clear and concise, avoiding any unnecessary or apologetic language.
\\- You are \textbf{strongly encouraged} to create useful and reusable lemmas to solve the problem.
\\- The lemmas should be as general as possible (generalizable), and be able to cover a large step in proofs (non-trivial).
Please ensure that your proof is well-organized and easy to follow, with each step building upon the previous one.

\textbf{Here are some useful skills for reference:}\\
\textbf{Useful skills 1:}
\vspace{-12pt}
\begin{lstlisting}[style=isabelle]
lemma square_diff_identity:
  fixes x y z :: real
  shows "(x - y)^2 + (y - z)^2 + (z - x)^2 = 2 * (x^2 + y^2 + z^2 - x * y - y * z - z * x)"
proof -
  have "(x - y)^2 + (y - z)^2 + (z - x)^2 = (x^2 - 2 * x * y + y^2) + (y^2 - 2 * y * z + z^2) + (z^2 - 2 * z * x + x^2)"
    by (simp add: power2_diff)
  also have "... = 2 * (x^2 + y^2 + z^2 - x * y - y * z - z * x)"
    by (simp add: algebra_simps)
  finally show ?thesis by auto
qed
\end{lstlisting}
[5 more useful skills...] \\ 

\textbf{Statement:}
Show that for positive integer n, 2 divides $4^n$. \\
\textbf{Informal Proof:}\\
Step 1: Since n is positive, we can find a natural number m where $m+1=n$. Then we can show that 2 divides $4^{m+1}$
\\
\textbf{Formal statement:}
\vspace{-12pt}
\begin{lstlisting}[style=isabelle]
theorem numbertheory_2dvd4expn:
  fixes n :: nat
  assumes h0 : "n \<noteq> 0"
  shows "(2::nat) dvd 4^n"
\end{lstlisting}
\vspace{-6pt}
\textbf{Formal proof:}
\vspace{-12pt}
\begin{lstlisting}[style=isabelle]
theory Scratch
  imports Complex_Main
begin
theorem numbertheory_2dvd4expn:
  fixes n :: nat
  assumes h0 : "n \<noteq> 0"
  shows "(2::nat) dvd 4^n"
proof -
  obtain m::nat where c0: "m+1=n"
    sledgehammer
  have "(2::nat) dvd 4^(m+1)" sledgehammer
  then show ?thesis unfolding c0 sledgehammer
qed
end
\end{lstlisting}
\vspace{-6pt}
[1 more formalization example...] \\

\textbf{Statement:}
What is the least possible value of $(xy-1)^2+(x+y)^2$ for real numbers $x$ and $y$?  Show that it is 1.
\\
\textbf{Informal proof:}\\
Step 1: Expand the expression \((xy-1)^2+(x+y)^2\) to obtain \(x^2+2xy+y^2+x^2y^2-2xy+1\).\\
Step 2: Simplify the expression derived in Step 1 to obtain \(x^2+y^2+x^2y^2+1\).\\
Step 3: Apply the Trivial Inequality, which states that all squares are nonnegative, to argue that the minimum value of the expression is 1.\\
Step 4: Show that the minimum value of 1 can be achieved when \(x=y=0\).\\
\textbf{Formal statement:}
\vspace{-12pt}
\begin{lstlisting}[style=isabelle]
theorem amc12a_2021_p7:
  fixes x y ::real
  shows "1 \<le> ((x * y) - 1)^2 + (x + y)^2"
\end{lstlisting}
\vspace{-6pt}
\Sepline
\textbf{ \ \\\small Output:\\\ \\}
\textbf{Formal proof:}
\vspace{-12pt}
\begin{lstlisting}[style=isabelle]
theory Scratch
  imports Complex_Main
begin
lemma am_gm:
  fixes x y :: real
  shows "x^2 + y^2 \<ge> 2 * x * y"
proof -
  have "(x - y)^2 \<ge> 0"
    by simp
  then have "x^2 - 2 * x * y + y^2 \<ge> 0"
    by (simp add: algebra_simps power2_diff)
  then have "x^2 + y^2 \<ge> 2 * x * y"
    by simp
  then show ?thesis 
    by simp
qed
theorem amc12a_2021_p7:
  fixes x y ::real
  shows "1 \<le> ((x * y) - 1)^2 + (x + y)^2"
  apply (auto simp:algebra_simps power2_eq_square)
  by (metis am_gm add.commute add.left_commute add_mono_thms_linordered_semiring(1) add_pos_nonneg less_add_same_cancel1 power2_sum power_mono zero_le_power2)
end
\end{lstlisting}

\end{tiny}
\end{tcolorbox}
\caption{Prompt example for formalization.}
\label{fig:formalization}
\end{figure}

\begin{figure}[H]
\begin{tcolorbox}[colback=mybrown!5!white,colframe=mybrown!75!black]
\begin{tiny}
\textbf{\small Input:\\\ \\}
\textbf{System message:}
As a mathematician familiar with Isabelle, your task is to provide a formal proof in response to a given formal statement.
Your proof should be structured and clearly written, meeting the following criteria:
\\- It can be verified by Isabelle.
\\- Please ensure that your proof is well-organized and easy to follow, with each step building upon the previous one.


\textbf{Formal statement:}
\vspace{-12pt}
\begin{lstlisting}[style=isabelle]
lemma power_eq_imp_eq:
  fixes a::real
  assumes "a>0" "a\<noteq>1" "a^m = a^n"
  shows "m = n"
\end{lstlisting}
\vspace{-6pt}
\textbf{Formal proof:}
\vspace{-12pt}
\begin{lstlisting}[style=isabelle]
theory Scratch
  imports Complex_Main
begin
lemma power_eq_imp_eq:
  fixes a::real
  assumes "a>0" "a\<noteq>1" "a^m = a^n"
  shows "m = n"
proof -
  have "a^(m-n) = 1"
    using assms(3) assms(2) by (metis assms(1) diff_is_0_eq diff_self_eq_0 le_numeral_extra(3) less_imp_diff_less linorder_le_less_linear nat_int_comparison(2) nle_le one_le_power order_le_less order_less_irrefl order_refl power_0 power_0_left power_decreasing_iff power_eq_0_iff power_inject_exp real_root_ge_1_iff real_root_pos_unique real_root_power zero_le_power)
  hence "m-n=0"
    by (smt (verit) assms(1) assms(2) assms(3) cancel_comm_monoid_add_class.diff_cancel diff_is_0_eq' power_decreasing_iff power_inject_exp verit_comp_simplify1(3))
  thus "m = n"
    by (smt (verit) assms(1) assms(2) assms(3) less_numeral_extra(3) nat_neq_iff power_inject_exp power_strict_decreasing_iff zero_less_diff)
qed
end
\end{lstlisting}
\vspace{-6pt}
[3 more request solving examples...] \\

\textbf{Formal statement:}\\
\vspace{-21pt}
\begin{lstlisting}[style=isabelle]
lemma exponent_properties:
  fixes a b :: real
  assumes "0 < a \<and> 0 < b"
  shows "a^n * a^m = a^(n + m) \<and> (a^n)^m = a^(n * m)"
\end{lstlisting}
\vspace{-6pt}
\Sepline
\textbf{ \ \\\small Output:\\\ \\}
\textbf{Formal proof:}
\vspace{-12pt}
\begin{lstlisting}[style=isabelle]
theory Scratch
  imports Complex_Main
begin
lemma exponent_properties:
  fixes a b :: real
  assumes "0 < a \<and> 0 < b"
  shows "a^n * a^m = a^(n + m) \<and> (a^n)^m = a^(n * m)"
proof
  show "a^n * a^m = a^(n + m)"
    by (simp add: assms(1) power_add)
next
  show "(a^n)^m = a^(n * m)"
    by (simp add: assms(1) power_mult)
qed
end
\end{lstlisting}

\end{tiny}
\end{tcolorbox}
\caption{Prompt example for solving request.}
\label{fig:solve_request}
\end{figure}

\begin{figure}[H]
\begin{tcolorbox}[colback=mybrown!5!white,colframe=mybrown!75!black]
\begin{tiny}
\textbf{\small Input:\\\ \\}
\textbf{System Message:}
As an expert mathematician who is proficient in Isabelle theorem proving, your task is to modify the given lemma, theorem, function, or definition given in the code to aid in solving one or more of the problems provided. You should accomplish this by [evolve description].
\textbf{Here are some reference problems:}\\
\textbf{Problem 1}:
\vspace{-12pt}
\begin{lstlisting}[style=isabelle]
theorem mathd_algebra_131:
  fixes a b :: real
    and f :: "real \<Rightarrow> real"
  assumes h0 : "\<And>x. f x = 2 * x^2 - 7 * x + 2"
    and h1 : "f a = 0"
    and h2 : "f b = 0"
    and h3 : "a \<noteq> b"
  shows "1 / (a-1) + 1 / (b-1) = -1"
\end{lstlisting}
\vspace{-6pt}
[3 more reference problems...] \\

\textbf{Skill to evolve:}
\vspace{-12pt}
\begin{lstlisting}[style=isabelle]
theory Scratch
  imports Complex_Main
begin
lemma mathd_numbertheory_001:
  fixes n :: nat
  shows "n mod 3 = 0 \<or> n mod 3 = 1 \<or> n mod 3 = 2"
by auto
end
\end{lstlisting}
\textbf{Evolved skill:}
\vspace{-12pt}
\begin{lstlisting}[style=isabelle]
theory Scratch
  imports Complex_Main
begin
lemma remainder_modulo:
  fixes n d :: nat
  assumes "d > 0"
  shows "n mod d \<in> {0..d-1}"
proof -
  have "n mod d < d" by (rule mod_less_divisor[OF assms(1)])
  moreover have "n mod d \<ge> 0" by simp
  ultimately show ?thesis by auto
qed
end
\end{lstlisting}
[2 more directional evolve examples] \\

\textbf{Skill to evolve:}
\vspace{-12pt}
\begin{lstlisting}[style=isabelle]
theory Scratch
  imports Complex_Main
begin
lemma divide_cross_mul:
  fixes a b c d :: real
  assumes "b \<noteq> 0"
    and "d \<noteq> 0"
    and "a / b = c / d"
  shows "a * d = b * c"
  using assms by (auto simp: field_simps)
end
\end{lstlisting}
\Sepline
\textbf{ \ \\\small Output:\\\ \\}
\textbf{Evolved skill:}
\vspace{-12pt}
\begin{lstlisting}[style=isabelle]
theory Scratch
  imports Complex_Main
begin
lemma divide_cross_mul_generalized:
  fixes a b c d x y :: real
  assumes "b \<noteq> 0"
    and "d \<noteq> 0"
    and "a / b = c / d"
    and "a = x * b"
    and "c = y * d"
  shows "x * d = y * b"
  using assms by (auto simp: field_simps)
end
\end{lstlisting}
\end{tiny}
\end{tcolorbox}
\caption{Prompt examples for directional evolve}
\label{fig:directional_evolve}
\end{figure}

\end{document}